\documentclass{bmvc2k}

\usepackage{microtype}
\usepackage{graphicx}
\usepackage{subfigure}
\usepackage{booktabs} 
\usepackage[utf8]{inputenc} 
\usepackage[T1]{fontenc}    
\usepackage{hyperref}       
\usepackage{url}            
\usepackage{amsfonts}       
\usepackage{nicefrac}       
\usepackage{microtype}      
\usepackage{xspace}
\usepackage{amsmath}
\usepackage{amssymb}
\usepackage{multirow}
\usepackage{bm}
\usepackage{cleveref}
\usepackage{todonotes}
\usepackage{xcolor}
\usepackage{cancel}

\title{Inductive Visual Localisation: Factorised Training for Superior Generalisation}

\addauthor{Ankush Gupta\\Andrea Vedaldi\\Andrew Zisserman}{{ankush,vedaldi,az}@robots.ox.ac.uk}{1}
\addinstitution{
 Visual Geometry Group\\
 Department of Engineering Science\\
 University of Oxford\\
}

\runninghead{Gupta, Vedaldi, Zisserman}{Inductive Visual Localisation}

\newcommand{\bx}{\mathbf{x}}
\newcommand{\by}{\mathbf{y}}

\newcommand{\bs}{\mathbf{s}}
\newcommand{\bc}{\mathbf{c}}

\newcommand{\bmm}{\mathbf{m}}

\newcommand{\suppmat}{appendix}

\def\eg{\emph{e.g}\bmvaOneDot} 
\def\ie{\emph{i.e}\bmvaOneDot}

\def\etal{\emph{et al}\bmvaOneDot}

\begin{document}
\maketitle
\maketitle

\begin{abstract}
End-to-end trained Recurrent Neural Networks (RNNs) have been successfully applied to numerous problems that require processing sequences, such as image captioning, machine translation, and text recognition. However, RNNs often struggle to generalise to sequences longer than the ones encountered during training. In this work, we propose to optimise neural networks explicitly for \emph{induction}. The idea is to first decompose the problem in a sequence of inductive steps and then to explicitly train the RNN to reproduce such steps. Generalisation is achieved as the RNN is not allowed to learn an arbitrary internal state; instead, it is tasked with mimicking the evolution of a valid state. In particular, the state is restricted to a spatial memory map that tracks parts of the input image which have been accounted for in previous steps. The RNN is trained for single inductive steps, where it produces updates to the memory in addition to the desired output. We evaluate our method on two different visual recognition problems involving visual sequences: (1) text spotting, \ie joint localisation and reading of text in images containing multiple lines (or a block) of text, and (2) sequential counting of objects in aerial images. We show that inductive training of recurrent models enhances their generalisation ability on challenging image datasets.
\end{abstract}

\begin{figure*}[t]
\centering
\includegraphics[width=\linewidth]{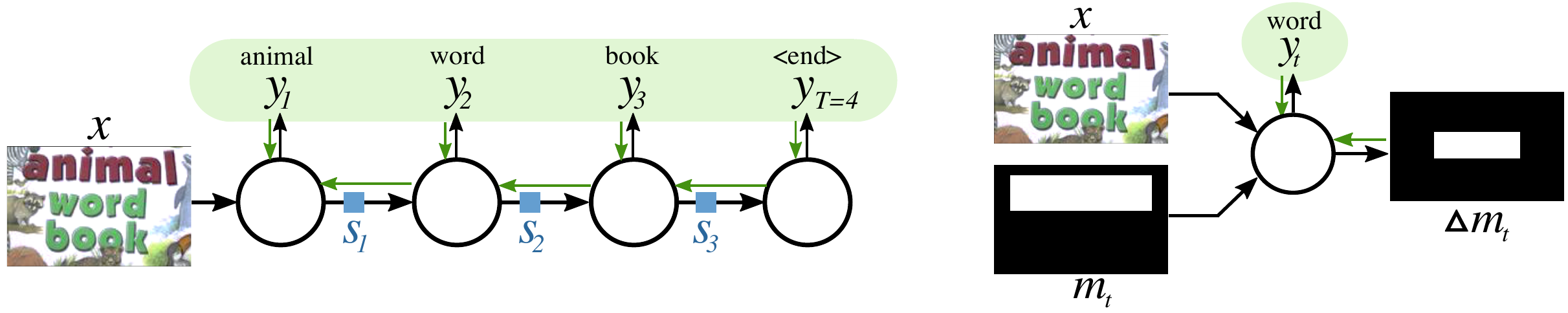}
\vskip -0.1in
\caption{\textbf{Inductive visual localisation.} A recurrent neural network (RNN) for sequence recognition is trained end-to-end on  sequences $(y_1,\hdots,y_T)$. The internal states ($\bs_t$) are learnt from gradients of losses summed over the entire sequence. Without structuring the state space, the recurrent unit may fail to learn the appropriate loop invariant, and thus may be unable to generalise to sequences longer than the length $T$ used for training. We address this problem by decomposing the end-to-end training procedure (left) into one-step inductive updates (right). We achieve this by restricting the recurrent state to a spatial memory map $\bmm_t$, which keeps track of the progress in processing the sequence. The recurrent network learns to incrementally predict, in addition to the output sequence $\by_t$, updates $\Delta\bmm_t$ to the memory. Using this inductive decomposition, our network can generalise well to sequences of length far greater than those in the training set.}\label{fig:model}
\end{figure*}

\section{Introduction}\label{sec:intro}
A key issue in sequence and program learning is to model long-term structure in the data. For example, in language modelling one has two choices: The first is to consider simple models such as character and word $n$-grams, which generalise well but fail to capture long-term correlations in the data. The second is to switch to models such as Recurrent Neural Networks (RNNs) that, in principle, can capture arbitrarily long correlations. In practice, however, RNNs are trained using back-propagation through time on sequences of limited length and may fail to generalise to longer sequences~\cite{Joulin15,Graves14a,Reed15,Hochreiter97}.

This is in contrast with the standard and very familiar notion of mathematical induction, which allows sequences to be analysed or generated ad infinitum. Many problems, such as counting objects or reading text, have an inherent inductive structure: all one needs to do is 1) count the current object or read the current character; and 2) move to the next object or character (or stop when finished). The two steps can then be iterated to process data of arbitrary length. However, as noticed by several authors~\cite{Joulin15,Graves14a,Reed15,Hochreiter97}, and confirmed in our experiments, RNNs fail to correctly repeat these steps beyond the number of times considered during training.

In this paper we marry the idea of induction to sequence processing using recurrent neural networks (RNN).
The inductive approach in this paper can be viewed as an application to the spatial domain of recent approaches for learning programs using recurrent and compositional networks~\cite{Cai17,Reed15}.

Our contributions are threefold: 1) we propose to train recurrent networks with the explicit notion of \emph{induction}, where the end-to-end training procedure is decomposed into inductive sub-steps. We show that RNNs trained on one-step inductive updates have superior generalisation ability. 2) We develop a recurrent module with inductive factorisation for recognising multiples lines of text in challenging scene text images, and outperform state-of-the-art methods by combining text localisation and recognition into a single architecture. 3) We apply our approach to sequential visual counting, and validate it on a challenging aerial image dataset, once again demonstrating improved generalisation capabilities.

The rest of the paper is organised as follows: in~\cref{sec:priorart} we first review related work; next, in~\cref{sec:model} we introduce our inductive decomposition approach; finally, in~\cref{s:exp} we present results on recognising multiple lines of text in images containing blocks of text, and on sequential visual object counting.

\section{Related Work}\label{sec:priorart}
\paragraph{Learning \& Composing Programs.}  Explicit decomposition into the repeated sub-tasks is related to the recent work on neural controllers, which interface with sub-programs or modules.
Neural Programmer-Interpreters~\cite{Reed15} present a general framework for dispatching functions using a program-stack, which Cai~\etal~\cite{Cai17} augment with the explicit notion of recursion, showing superior generalisation. Zaremba~\etal~\cite{Zaremba16}, learn controllers for operators acting on coarse 2D grids of discrete symbols, while our method works on dense pixel-grids.
\vspace{-5mm}%
\paragraph{RNNs for Scene Text Recognition}
Scene text recognition is a well-studied problem, with roots in optical and handwritten character recognition~\cite{Lecun98}. Traditional methods focused on single character recognition~\cite{Campos09,Yao14} with explicit language models (e.g., n-grams or lexicons) to form words or sentences~\cite{Wang11,Lee14, Alsharif14,Jaderberg14,Wang12,Mishra12,Shi13,Novikova12}. More recently, the word-level text recognition has been explored extensively~\cite{Goodfellow14,Jaderberg14c,Jaderberg15a,Poznanski16}, primarily using Convolutional Neural Networks (CNNs)~\cite{Lecun98} to encode images, and RNNs as the decoders~\cite{Su14,He16DTRN,ShiBY15,Lee16,Shi16}. A key component of the RNN decoders
is soft-attention~\cite{Bahdanau15} which iteratively pools image features at each step of the recurrence~\cite{Lee16,Shi16}.~\cite{Bluche16} extend this attention to 2D feature-maps for recognising multiple-lines of handwritten text.
\vspace{-5mm}%
\paragraph{RNNs for Visual Object Counting.} Counting objects in images~\cite{Arteta14} has numerous applications \eg, histological analysis of microscopy images~\cite{Arteta12}, parsing medical scans, and population studies from aerial imagery~\cite{Arteta16}. Sequential counting has been shown to be the primary method of counting in humans~\cite{dehaene1994dissociable}, and was explored with convolutional-RNNs in~\cite{romera2016recurrent}. More recently,~\cite{Zhang2017fcn} combine the object density based regression methods~\cite{Lempitsky10b} with iterative counting using recurrent fully-convolutional networks.
We factorise the end-to-end training of such iterative counting methods, and achieve superior generalisation.

\section{Method}\label{sec:model}
In order to generalise correctly to sequences of arbitrary lengths, an iterative algorithm
must maintain a suitable invariant.
For example, an algorithm that counts objects via enumeration maintains as invariant the list of objects visited so far, which must contain no repetition.
In order to maintain this invariant, the algorithm must visit at each step a ``new'' object, or terminate if no more objects are available.
However, RNNs are trained without any explicit
constraints on the structure of their hidden state, and may not learn such an invariant correctly.
For example, a list of objects has no a-priori limitation on its size. While for a human this is obvious, an RNN may be unable to understand it as it cannot experience unbounded lists during training.

In order to address this issue, we design an RNN to update a state that, by construction, has a universal step-independent validity. In particular, we
restrict the recurrent state to a \emph{spatial memory map}
which keeps track of the parts of the input image which have already been explored.
At each step, the model is conditioned on this memory map (in addition to the input image),
and predicts as output a token for the sequence, as well as an update to the spatial memory.
Such updates for the counting example above amount to adding one more object to the list of visited objects.
This is analogous to taking the \emph{inductive step} in mathematical induction.

The rest of the section describes our model in detail. In~\cref{s:edm} we discuss encoder-decoder RNNs~\cite{Cho14,Sutskever14,Vinyals15} enhanced with soft-attention~\cite{Bahdanau15, Xu15} as these are suitable for modelling sequences in images. In~\cref{s:inductive} we describe our inductive decomposition for these recurrent models. Finally, in~\cref{s:training} we detail training and inference.

\subsection{Encoder-Decoder Models}\label{s:edm}
Let $\bx \in \mathcal{X}=\mathbb{R}^{H\times W\times C}$ be an image, where $H, W$ and $C$ are its height, width and number of color channels. Furthermore, let $\by = (y_1,\hdots,y_T) \in \mathcal{Y}^T$ be the corresponding sequence label, where $T$ is the sequence length.
Encoder-decoder methods model the conditional probability $p(\by | \bx)$ as a product of conditionals for the next token $y_{t+1}$ given a context vector $\bc_t$ and the previously-predicted tokens $y_{1:t}$:
\begin{equation}\label{e:joint}
	p(\by | \bx) = \prod_{t=0}^{T-1} p(y_{t+1} | y_{1:t}, \bc_t)
\end{equation}
These conditional probabilities are modelled using an RNN $\Phi$. We consider in particular an LSTM~\cite{Hochreiter97} with hidden state $\bs_t$, and write:
\begin{equation*}
	\Big(p(y_{t+1} | y_{1:t}, \bc_t),\hspace{2mm} \bs_{t+1}\Big) = \Phi(\bs_t, \bc_t, y_t).
\end{equation*}
The context vector $\bc_t$ injects into the model information extracted from the input image. Context can be kept constant for all steps, or can be dynamically focused on different parts of the image using an attention mechanism.
In the first case, exemplified by sequence-to-sequence models~\cite{Vinyals15}, the context is extracted\footnote{Since in this case no attention is used, the CNN is usually configured so that $H'=W'=1$.}  by a Convolutional Neural Network (CNN)~\cite{Lecun89} $\bc_t = \bc = \Psi(\bx) \in \mathbb{R}^{H'\times W'\times C'}$. In the second case, exemplified by~\cite{Bahdanau15}, the context vector is computed at each step via attention by reweighing the CNN output:
\begin{align}~\label{e:attn}
	\bc_t &= \sum_{i \in H'}\sum_{j \in W'} \alpha_{ij} \Psi(\bx)_{ij}, \\
	\alpha_{ij} &= \frac{\exp(v_{ij})}{\sum_{i'}\sum_{j'} \exp(v_{i'j'})}, \label{eq:alpha}\\
	v_{ij} &= w^T\tanh(W\bs_t + W'\Psi(\bx)_{ij} + b),
\end{align}
where, $v_{ij} \in \mathbb{R}$ is the unnormalised attention score and $w, W, W',b$ are learnable parameters.

The model is trained end-to-end to maximize the log of the posterior probability~\eqref{e:joint} averaging over example (image $\bx$, sequence $\by$) pairs.

\subsection{Inductive Decomposition}\label{s:inductive}
The RNN models discussed in the previous section may fail to learn a correct inductive decomposition of the problem, and thus fail to generalise properly to sequences of arbitrary length.
We propose to address this problem in a simple and yet effective manner: rather than allowing the RNN to learn its own state space, we specify a suitable state space a priori, and train the RNN to make use of it.
In more detail, we set the RNN state to be a spatial memory $\bs_t=\bmm_t$ containing a mask covering all the visual objects that have  been accounted for up to time $t$. In this manner, the content of the memory can be derived from the ground-truth data annotations and the step number $t$. Furthermore, the RNN does not need to learn a new state space from scratch, but only how to generate $\bmm_{t+1}$ from $\bmm_{t}$; for this, training can focus on learning single-step predictions, from $t$ to $t+1$, rather than whole-sequence predictions.

The spatial memory $\bmm \in \mathbb{R}^{H\times W}$
is implemented as a single 2D map of the same dimensions as the image $\bx$.
At each step, the model predicts $y_t$, as well as an update $\Delta \bmm_t$ to the memory.
In this work, we focus on sequence prediction tasks where each token in the
sequence corresponds to a 2D location in the image.
Hence, $\Delta \bmm_t$ is trained to encode the 2D location in the image associated with $y_t$.

Predictions at each step are conditioned on the context vector $\bc_t$,
as well as the memory $\bmm_t$:
\begin{equation}
	\Big(p(y_{t+1} | y_{1:t}, \bc_t),\hspace{2mm} \Delta \bmm_{t+1}\Big) = \Phi(\bc_t, \bmm_t).
\end{equation}
In practice, at each step, the memory is concatenated with the image to obtain the encoded
representation $\Psi([\bx || \bmm])$ instead of being fed into $\Phi$ directly;
$\bc_t$ is obtained from $\Psi$ as detailed in~\cref{s:edm}.
The memory is initialised to all zeros, \ie, $\bmm_{t=0} = 0^{H\times W}$, and is updated after
each step as:
\begin{equation}\label{e:mask}
	\bmm_{t+1} = \bmm_t + \Delta\bmm_t.
\end{equation}
The exact architecture of $\Phi$ and the location representation in $\bmm_t$
are application dependent, and examples are detailed in \cref{s:text-model,s:count-model}.

\subsection{Training and Inference}\label{s:training}
\begin{figure*}[t]
\centering
\includegraphics[width=\linewidth]{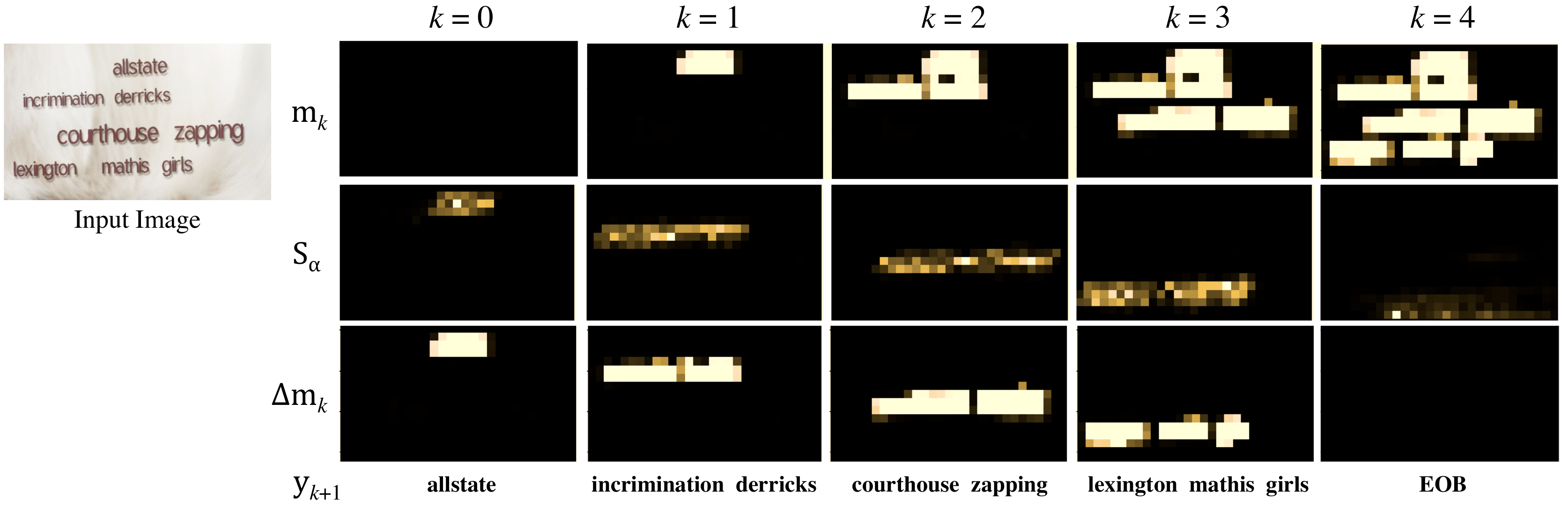}
\vskip -0.1in
\caption{\textbf{Multi-line text recognition inference visualisation.}  Decoding procedure for an example image containing four lines. Inference runs for five steps, predicting the line characters ($y_k$) in the first four steps, and indicating the end-of-block (\texttt{EOB}) in the fifth step. The memory-map is initialised (at $k{=}0$) to all zeros, and is iteratively updated by adding in the predicted update $\Delta\bmm_k$, regressed from the
accumulated attention maps $S_\alpha$.}
\label{fig:inference-vis}
\end{figure*}

\paragraph{Training.} The model is trained for one-step predictions, where each
training sample is a tuple --- $(\bx, y_t, \bmm_t, \bmm_{t+1})$: image, token at time $t$,
ground-truth accumulated location maps $\bmm_t$ and $\bmm_{t+1}$.
The model is optimised through stochastic gradient descent (SGD) to minimise the
following loss:
\begin{equation}
	-\log p(y_t | \bx_t, \bmm_t) + \gamma ||\bmm_t + \Delta\bmm_t - \bmm_{t+1}||_2^2
\end{equation}
where, $\gamma>0$ balances the terms. The first term maximises the probability of the
correct token $y_t$, while the second is a pixel-wise reconstruction loss for
the predicted spatial memory update $\Delta\bmm_t$.
The sequence label for the final step $(y_T)$ indicates the end-of-sequence,
\eg through an additional class in the output labels; it is used for terminating the inference loop.
\vspace{-5mm}
\paragraph{Optimisation.} All model parameters are initialised randomly (sampled from a gaussian with 0.01 standard deviation). The model is trained with SGD using the AdaDelta optimiser~\cite{Zeiler12}.
\vspace{-5mm}
\paragraph{Inference.} The memory is initialised to all zeros. At each step, the memory
is concatenated with the image and fed through the image encoder to get the encoded-representation
$\Psi([\bx || \bmm])$. Then, the log-probabilities for $y_t$, and memory updates $\Delta\bmm_t$
are regressed from the recurrent module $\Phi$. The memory is updated per~\cref{e:mask},
and fed into the model iteratively, until the end-of-sequence is predicted. An example is shown in~\cref{fig:inference-vis}.

\section{Experiments}\label{s:exp}
We evaluate our method on two different tasks --
 in \cref{s:text} we present results on recognising lines of text in
images containing a block or multiple lines of text, and
in \cref{s:count} we explore counting objects in images through enumeration.
We demonstrate superior generalisation ability of recurrent modules
when trained with inductive factorisation, and outperform state-of-the-art
methods on a text-spotting task.
\subsection{Recognising Multiple Lines of Text}\label{s:text}
We apply our model to the task of spotting, \ie joint localisation
and recognition, of text in images containing multiple lines (or a block)
of text. We proceed in two steps: first, using synthetically generated
text-block images, we show superior generalisation ability of the model
trained with inductive factorisation; second, we fine-tune this text-block
spotting model trained on synthetic data, on real text data extracted
from the ICDAR~2013~\cite{Karatzas13} benchmark, and outperform
the state-of-the-art word-level text spotting method.

\subsubsection{Model Details}\label{s:text-model}
\begin{figure*}[t]
 \centering
  \begin{minipage}[t]{\linewidth}
	\centering
	\includegraphics[width=\linewidth]{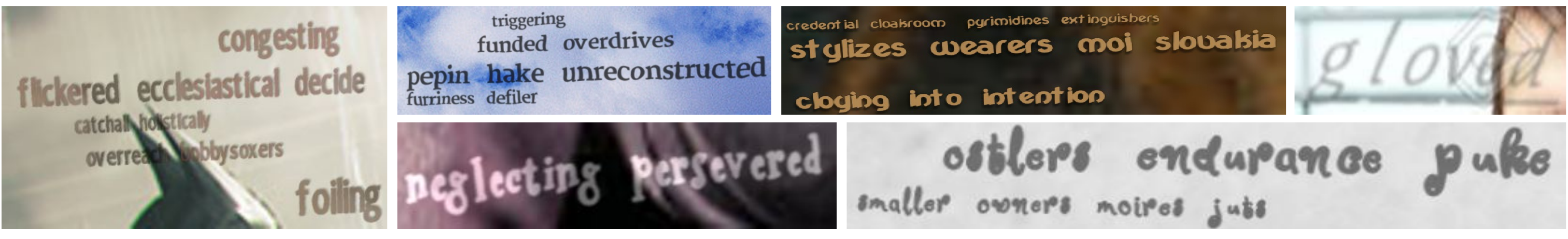}\vspace{-1em}
  \end{minipage}

  \begin{minipage}[b]{0.95\linewidth}
    	\resizebox{\linewidth}{!}{
        \begin{tabular}{ llrrrrrrrrrr }%
    \toprule
      &\# lines$\rightarrow$ & 1 & 2 & 3 & 4 & 5 & 6 & 7 & 8 & 9 & 10 \\
      \midrule%
\multirow{3}{*}{end-to-end}       &precision & 66.69 & 63.97 & 59.23 & 53.70& - & - & - & - & - & - \\
      &recall & 69.27 & 65.50 & 56.52 & 39.14& - & - & - & - & - & - \\
      &ED & 15.91  & 17.81 & 25.08 & 43.78& - & - & - & - & - & - \\\hline
\multirow{3}{*}{inductive}       &precision & 85.13 & 84.79 & 85.57 & 87.25 & 87.32 & 86.11 & 85.41 & 85.51 & 84.57 & 84.41 \\
      &recall & 84.89 & 84.74 & 85.32 & 86.99 & 87.22 & 85.91 & 84.43 & 84.03 & 80.47 & 76.80 \\
      &ED & 6.76  & 7.79  & 7.09  & 6.29  & 5.77  & 6.96  & 8.77  & 9.11  & 13.18 & 17.23\\
      \bottomrule
    \end{tabular}}
    \vskip 0.02in
    	\caption{\textbf{Synthetic Text Blocks results.} (top) Samples with different number of text lines from the Synthetic Text Blocks test set. (bottom) Word-level precision, recall, and character-level normalised edit-distance (ED) are reported (all in \%).}
    \label{fig:synth}
\end{minipage}
\end{figure*}

Given an image $\bx$ containing multiple lines of text (or a text-block; see examples in~\cref{fig:synth}),
the corresponding sequence label $\by$ is a sequence of characters
in lexicographic order (\ie left-to-right, first-to-last line).
We factorise the problem of spotting text in text-blocks at the level of lines,
\ie a ``token'' $y_k$ (in~\cref{s:inductive}), corresponds to the $k^{th}$ line.
Hence, $y_k$ itself is a sequence of $n$ characters $\{y^k_1,\hdots,y^k_n\}$ in the
$k^{th}$ line. At each inductive step, one full-line is recognised.
\vspace{-5mm}%
\paragraph{Spatial Memory Representation.} The spatial memory $\bmm_k$ at step $k$
represents the location of first $k$ lines which have been recognised so far,
by setting the pixel-values inside the corresponding line-level bounding-boxes
to 1 (the background is 0). The memory updates $\Delta\bmm_k$ correspond
to the location of $k^{th}$ line (see~\cref{fig:inference-vis}).
\vspace{-5mm}%
\paragraph{Image Encoder ($\Psi$).}
We employ the fully-convolutional DRN-C-26 Dilated Residual Network~\cite{Yu17} as the image-encoder, which consists of six residual blocks; the encoder downsamples the image by a factor of 8, and has a stride of 32 (details in the reference, and \suppmat).
Hence, an input image, concatenated with memory-map ($\bmm$) of dimensions $H\times W\times 4$ is encoded as a feature-map of dimensions $\lceil\frac{H}{8}\rceil\times\lceil\frac{W}{8}\rceil\times 512$.
\vspace{-5mm}%
\paragraph{Recurrent Module ($\Phi$).} We use an  LSTM-RNN with soft-attention
over the convolutional features as the line-level character decoder. The state-size is set to 1024,
while the attention-embedding dimension is set to 512.
The attention weights ($\alpha$ in~\cref{eq:alpha}) corresponding to all the predicted
characters in the current line are summed up; this produces an approximate localisation ($S_\alpha$)
of the line (second row in~\cref{fig:inference-vis}). $S_\alpha$ is concatenated with the image-representation
$\Psi$, and convolved with a stack of 2 convolutional+ReLU~\cite{Nair10} layers (128 filters each), and a final $1{\times}1$ convolutional layer to produce the memory-update ($\Delta\bmm$).

\subsubsection{Synthetic Text Blocks}\label{s:synth-text}
Following the success of synthetic data in text-spotting~\cite{Wang12,Jaderberg14c,Gupta16},
we test the generalisation to number of lines beyond those present in the training set,
on synthetically generated text-block images (\cref{fig:synth}).
Using synthetic data enables this study, as it is difficult to collect real-world images
of text with a large number of lines.
\vspace{-5mm}%
\paragraph{Dataset.} The training set consists of text-block containing 3--5 lines,
while the test contains 1--10 lines, with 500000 samples for each number of lines.
To generate a synthetic text-block image for a given number of lines,
the following procedure is followed: a random number of words (3--5 per line) are
selected from a lexicon of approximately 90k words~\cite{Jaderberg14c}.
Then the text-lines are randomly aligned (left, centre, right),
resized to potentially different heights (within the same block),
separated by random amounts of line-spacing,
transformed with a small perspective or affine transformation,
and finally rendered against a randomly chosen background image,
with a font chosen from over 1200 fonts.
\Cref{fig:synth} shows some samples generated through this procedure.
\vspace{-5mm}%
\paragraph{Evaluation Metrics.}  We report the word-level precision and recall,
computed as the intersection of the predicted words and the ground-truth words,
normalised by the total number of predicted or ground-truth words respectively,
in addition to the normalised edit-distance.
\vspace{-5mm}%
\paragraph{Results.}  The results are summarised in \cref{fig:synth}, and \cref{fig:inference-vis} visualises the predicted updates to the mask on a test image.  We note consistent levels of precision and recall, even when testing with twice the number of lines than in the training set.
We also trained a RNN model (with identical $\Psi, \Phi$) end-to-end
without any inductive factorisation. It suffered from two difficulties:
(1) block-level end-to-end training did not converge for more than three lines;
and (2) the model did not generalise to more than three lines: note,
the steep fall in the recall rates, and increase in the edit-distance for four lines.
\subsubsection{Real Text Blocks -- ICDAR-2013}
\begin{figure*}[t]
 \centering
  \begin{minipage}[t]{\linewidth}
	\centering
	\includegraphics[width=\linewidth]{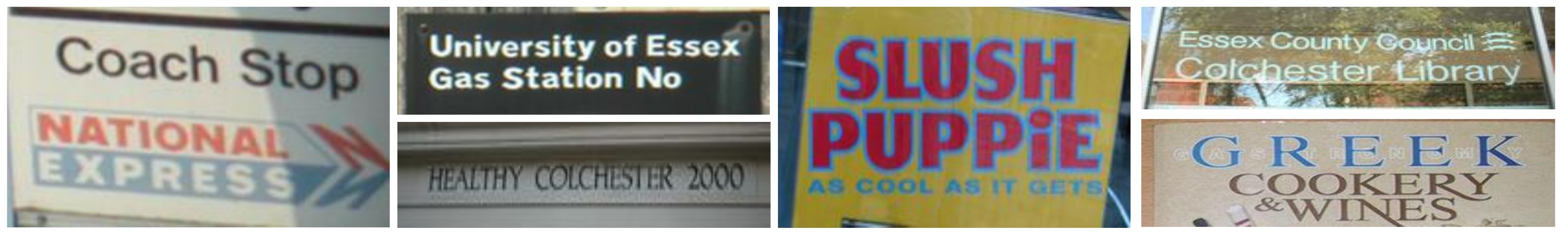}\vspace{-1em}
  \end{minipage}

  \begin{minipage}[b]{0.95\linewidth}
    	\resizebox{\linewidth}{!}{
    \begin{tabular}{ lcrrrrrr }%
    \toprule
      \# lines$\rightarrow$ && 1 & 2 & 3 & 4 & 5  & overall\\
      \hline%
      \# blocks & & 169 & 79 & 29 & 6 & 3 & 117\\
      \# words & & 270 & 267 & 136 & 34 & 40 & 477\\
      \midrule
\multirow{3}{*}{He~\etal\cite{He17} and  Jaderberg~\etal\cite{Jaderberg14c}} & P & \textbf{96.15}   & \textbf{95.59}          &  \textbf{94.59}     & \textbf{98.92}          & \textbf{98.74}          & \textbf{95.94} \\
                      & R                                      & 61.40   & 72.22          & 76.09      & 64.71          & 72.50          & 68.58          \\
                      & F                                      & 74.95   &  82.28          & 84.34      & \textbf{78.23}          & 83.68          & 79.98          \\
                      \midrule
\multirow{3}{*}{ours} & P & 80.31 & 84.09        & 86.76 & 72.97          & 87.18         & 82.24        \\
                      & R & \textbf{77.04} & \textbf{83.15} & \textbf{86.76} & \textbf{79.41} & \textbf{85.00}   & \textbf{81.06}  \\
                      & F & \textbf{78.64} & \textbf{83.62} & \textbf{86.76}  & 76.06& \textbf{86.08} & \textbf{81.65}  \\
                      \bottomrule
    \end{tabular}}
   \vspace{-3mm}
    \caption{\textbf{ICDAR-2013 Text Blocks results.} (image) Samples with different number of text lines from the ICDAR-2013 Text Blocks test set. (table) Number of block-images with the given number of text lines (\emph{\# blocks}), the total number of words in these images (\emph{\# words}), and the word-level precision (\emph{P}), recall (\emph{R}), and F-score (\emph{F}) (all in \%) are reported.}
    \label{fig:icdar}
\end{minipage}
\end{figure*}

We fine-tune the inductive block-parser trained on synthetic data
on text-blocks extracted from the ICDAR 2013 Focussed Scene Text~\cite{Karatzas13} dataset,
and compare with state-of-the-art word-level methods
\vspace{-5mm}%
\paragraph{Dataset.} ICDAR 2013 is a dataset of 229 training,
and 233 test scene images containing text, with word-level bounding
boxes and text-string annotations.
As our model is tailored for recognising text in blocks,
we extract images of text-blocks, \ie images containing multiple lines of text
from the dataset:
first, text lines are formed by linking together pairs of words with relative distance at most
half, and three times the minimum of their heights, along the vertical and horizontal directions respectively;
then, the bounding boxes of these lines are doubled along the height, and those with
area of intersection at least a third of the maximum of their areas are merged into text-blocks.
Some samples are visualised, and the number of
blocks obtained and other statistics are given in \cref{fig:icdar}.
\vspace{-5mm}%
\paragraph{Baseline word level model.}  We combine the
word localisation model of  He~\etal~\cite{He17} (88\% F-score on ICDAR13 Focussed Scene localisation),
with the strong lexicon based word-level recognition network of~\cite{Jaderberg14c} (90.8\% on ICDAR13 cropped word recognition).\footnote{Implementations were obtained from the authors.}
This combination of state-of-the-art models for localisation and recognition, provides a strong baseline to compare our joint model against. To minimise the discrepancy between the test and training setting we run the detector on full scene images and then using the detected word locations, associate them with the text-blocks
used in this experiment. Note, we also ran the baseline detector on block images but this lead to worse results: word-spotting F-score = 78.83\% (block-images) vs.\  79.98\% (full-scene images).
\vspace{-5mm}%
\paragraph{Evaluation.}
As is standard in the benchmark, we report the word-level recall, precision and the F-score.
For fair comparison with the lexicon-based word-recognition model~\cite{Jaderberg14c}, we use the same lexicon of 90k words to constrain the predictions of our model.
\vspace{-5mm}%
\paragraph{Results.} \Cref{fig:icdar} summarises the results.
We note that our inductive block parser, consistently achieves a higher
recall and F-score (except for 4-lines) across all lines.
Our method combines the stages of both localisation and recognition,
and hence avoids downstream error propagation, achieving greater recall.
The higher precision of the baseline is due to the detector only producing
high-confidence detections.

\subsection{Counting by Enumeration}\label{s:count}
We further test improvement in generalisation ability induced by inductive training in a different
setting: counting objects in images through enumeration, where
a recurrent module acts as an  enumerator, counting one object
in each step, and terminating when done. We test on two
datasets: first, images containing randomly coloured shapes,
and second: aerial images of airplanes extracted from the recently
introduced DOTA dataset~\cite{xia2018dota}.

\subsubsection{Model Details}\label{s:count-model}
\begin{figure*}[t]
\centering
\includegraphics[width=0.6\linewidth]{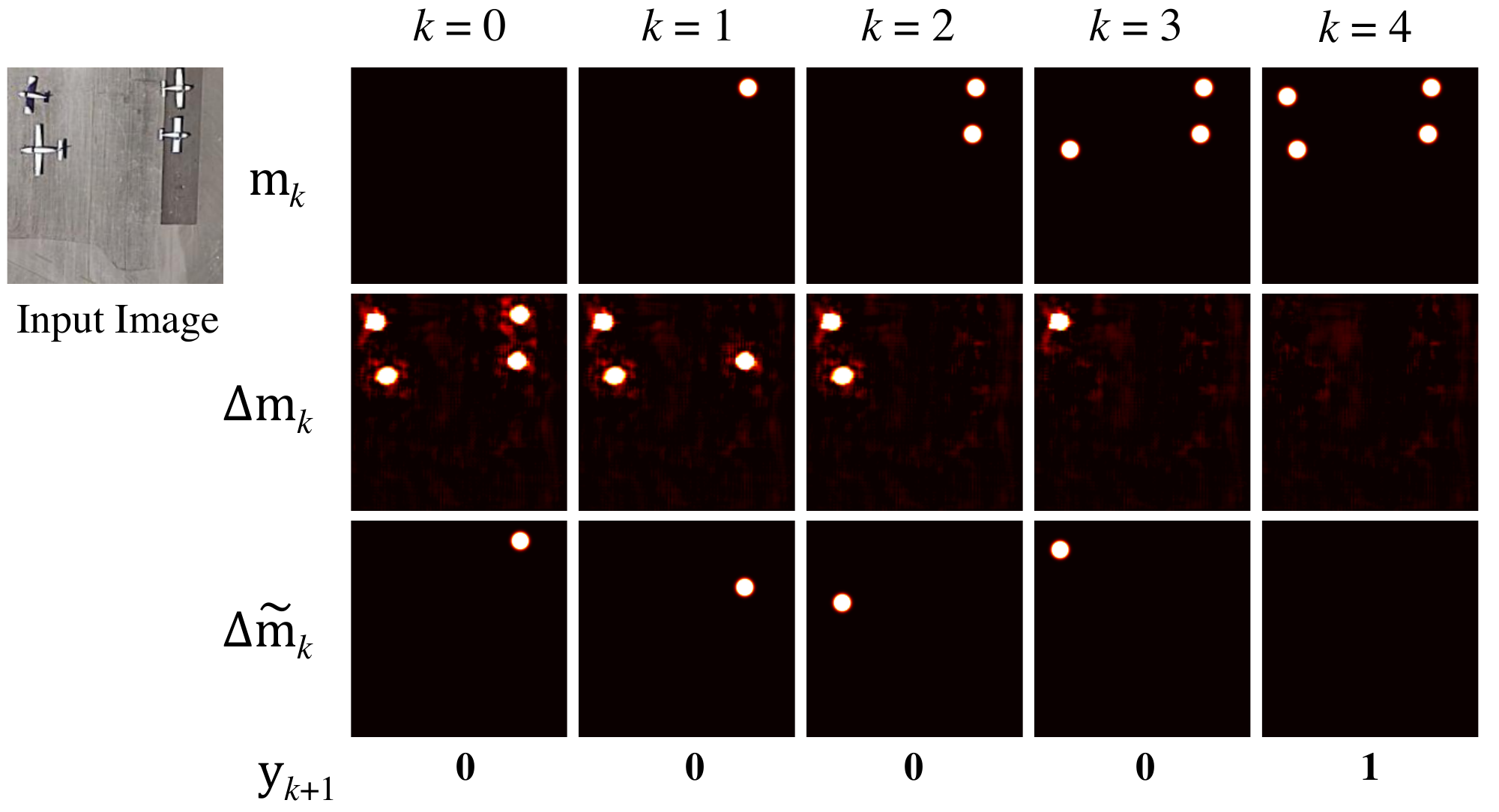}
\vskip -0.1in
\caption{\textbf{Enumerative counting visualisation.}  Decoding procedure for an example image containing four airplanes. Inference runs for five steps, indicating the end-of-sequence ($y_k = 1$) in the fifth step. At each step, the memory update ($\Delta m_k$) regresses peaks at the location of \emph{all} the remaining objects, out of which one is randomly picked ($\Delta\tilde{m}_k$) to update the memory-map for the next step.}
\label{fig:inference-count-vis}
\end{figure*}

In each inductive step one object is counted, and a corresponding label of `0'
is produced; the enumeration terminates when all the objects have been accounted for,
producing `1' as the output, as in~\cite{romera2016recurrent}.
Hence, for a given image $\bx$ containing multiple ($=N \geq 0$)
objects of interest, the corresponding sequence label is $(0,\hdots,0,1)$, \ie
a sequence containing $N$ zeros, and one 1.
\vspace{-5mm}%
\paragraph{Spatial Memory Representation.} The spatial memory $\bmm_k$ at step $k$
represents the location of first $k$ objects which have been enumerated, by placing a
small gaussian peak at their centre. Due to lack of any natural order for counting,
it is difficult to assign a particular object to a specific enumeration step.
Hence, the memory update $\Delta\bmm_k$ at the $k^{th}$ step regresses
gaussian peaks at the locations of \emph{all} the remaining objects.
The memory is updated with the location of one of the remaining objects (selected randomly);
~\cref{fig:inference-count-vis} visualises the inference steps for a sample image.
\vspace{-5mm}%
\paragraph{Image Encoder ($\Psi$).}
A fully-convolutional Dilated Residual Network~\cite{Yu17}, consisting of three residual-blocks
each with two pre-activation residual units~\cite{he2016identity}, is used to encode the images (detailed architecture in \suppmat). The images are not downsampled, hence, the feature-maps retain the original
dimensions of the input.
\vspace{-5mm}%
\paragraph{Recurrent Module ($\Phi$).}
The memory updates $\Delta\bmm_k$ are regressed from the image-features,
using a $1{\times}1$ convolutional layer.
The binary valued token at each step $y_k$ is predicted as:
$p(y_k=1) = \operatorname{Sigmoid}\left(w\cdot\operatorname{MaxPool}(\Delta\bmm_k) + b\right)$.

\subsubsection{Datasets}
\begin{figure*}[t]
 \centering
  \begin{minipage}[t]{\linewidth}
	\centering
	\includegraphics[width=\linewidth]{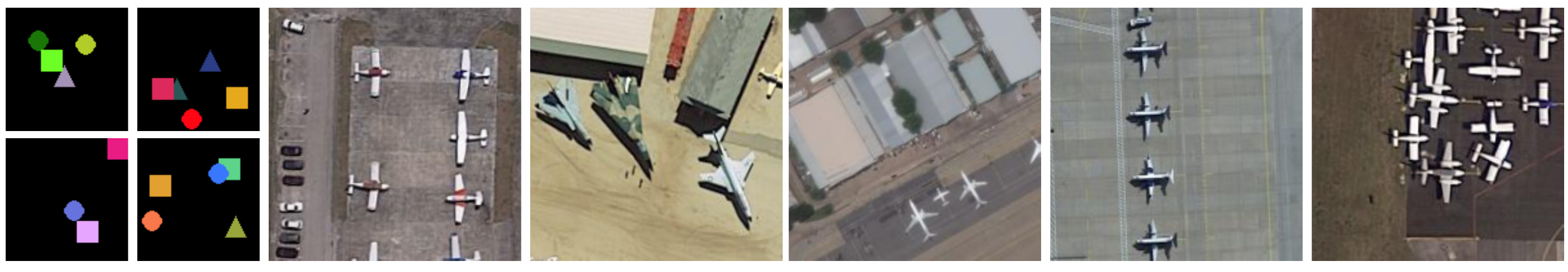}\vspace{-1em}
  \end{minipage}

  \begin{minipage}[b]{1.0\linewidth}
    	\resizebox{\linewidth}{!}{
\begin{tabular}{@{}llcccccccc@{}}
\toprule
\multicolumn{1}{c}{\multirow{2}{*}{Dataset}} & \multicolumn{1}{c}{\multirow{2}{*}{Model}} & \multicolumn{8}{c}{Number of Objects}                         \\ \cmidrule(l){3-10}
\multicolumn{1}{c}{}                         & \multicolumn{1}{c}{}                       & 3     & 4     & 5     & 6     & 7     & 8     & 9     & 10    \\ \midrule
\multirow{2}{*}{Coloured Shapes}             & end-to-end                                 & 99.53 & 99.53 & 98.93 & 0     & 0     & 0     & 0     & 0     \\
                                             & inductive                                  & 100   & 99.89 & 99.52 & 98.93 & 97.18 & 98.47 & 95.48 & 95.45 \\ \midrule
\multirow{2}{*}{DOTA}                        & end-to-end                                 & 82.00 & 70.50 & 74.80 & 0     & 0     & 0     & 0     & 0     \\
                                             & inductive                                  & 82.50 & 79.00 & 75.50 & 72.50 & 69.00 & 43.81 & 32.21 & 29.20 \\ \bottomrule
\end{tabular}
    }
   \vspace{-3mm}
    \caption{\textbf{Enumerative counting results.} (top) Samples from the Coloured Shapes and DOTA Airplane datasets containing different number of objects. (bottom) Accuracy of enumerative counting with and without inductive training (all in \%).}
    \label{fig:count}
\end{minipage}
\end{figure*}

We evaluate on two datasets described below. Images of size $128{\times}128$
were used for both the datasets. The training sets consisted of images containing $\{3,4,5\}$
objects; to test for generalisation beyond training sequence lengths, the test set included 3--10 objects.
\Cref{fig:count} visualises some samples from the datasets.
\vspace{-5mm}%
\paragraph{Coloured Shapes.}  We start with a procedurally generated synthetic dataset,
which consists of shapes with random colour, position and type (circle, triangle, or square) placed on the canvas. This dataset provides a simplified setting for analysis without confounding complexity.
The training set consists of 10k images, while the test set consists of 200 images for each count.
\vspace{-5mm}%
\paragraph{DOTA Airplanes.} DOTA~\cite{xia2018dota} is a recently introduced
image dataset of high-resolution aerial images, where 15 object categories have
been labelled with oriented bounding-boxes.
We extract image crops from the DOTA dataset which consist of
airplanes. The training set consists of 20k crops extracted from 100
images from the dataset's training set,
while the test set consists of 200 image crops for each object count (ranging from 3 to 10) extracted
from 70 images from the dataset's validation set.

\subsubsection{Results}
We compare our inductive model against a soft-attention LSTM-RNN trained end-to-end
with the same image encoder $\Psi$. We report the mean accuracy of prediction,
where a test image is evaluated as correct if the predicted count matches the
ground-truth count exactly.~\Cref{fig:count} summarises the results on both
the datasets for the two models. On both the datasets, end-to-end
trained RNN fails to generalise to object counts beyond those in the training
set ($>5$), while the one with inductive training does not fail catastrophically
at higher counts. Lower accuracy of the inductive model at higher counts
can be attributed to crowding of objects, which is not seen in the training set (\eg~\cref{fig:count} rightmost image).

\section{Conclusions}\label{sec:conc}
While RNNs may seem a perfect match for problems with an inductive structure, these networks fail to learn appropriate invariants to allow recursion to extend beyond what is encountered in the training data.
We have shown how to repurpose standard RNN models to restrict the recurrent state to a suitable state-representation -- which in our application are the image locations corresponding to the predictions at each step -- where the correct invariant can be enforced.
The result is an iterative visual parsing architecture which generalises well-beyond the training sequence lengths. This idea can be extended to visual problems with a tail-recursive structure, from object tracking to boundary and line tracing.

\paragraph{Acknowledgements.}
We thank Triantafyllos Afouras for proofreading.
Financial support was provided by the UK EPSRC CDT in Autonomous Intelligent Machines and Systems Grant EP/L015987/2, EPSRC Programme Grant Seebibyte EP/M013774/1, and the Clarendon Fund scholarship.

\bibliographystyle{ieee}
\bibliography{shortstrings,my,vgg_other,vgg_local}
\clearpage
\appendix\section*{Appendix}
In~\cref{sec:encoder} we first give detailed architecture of the image-encoder $\Psi$ used for text-recognition in multiple lines (section 4.1 in paper), and for visual object counting (section 4.2). Next, in~\cref{s:toy} we present results on a synthetic shapes dataset for the recognition task, similar to the one used counting; this was excluded from the paper due to lack of space. 
\section{Image Encoder Architecture ($\Psi$)} \label{sec:encoder}
Our image encoder is based on the Dilated Residual Network~\cite{Yu17}.
We give details of the architecture of the encoder used for text-recognition and counting respectively.

\subsection{Text Recognition Encoder}
The image encoder is based on the DRN-C-26 network of~\cite{Yu17}.
The network if fully-convolutional, downsamples the input by a factor of 8, and has a stride of 32.
The layer-level details are as following (top is first layer):\\

\noindent $\text{Conv}{-}5{\times}5{-}\text{F}16{-}\text{D}1$\\
$\text{Res}{-}3{\times}3{-}\text{F}16{-}\text{D}1{-}\text{S}2$\\
$\text{Res}{-}3{\times}3{-}\text{F}32{-}\text{D}1{-}\text{S}2$\\
$\text{Res}{-}3{\times}3{-}\text{F}64{-}\text{D}1$\\
$\text{Res}{-}3{\times}3{-}\text{F}64{-}\text{D}1{-}\text{S}2$\\
$\text{Res}{-}3{\times}3{-}\text{F}128{-}\text{D}1$\\
$\text{Res}{-}3{\times}3{-}\text{F}128{-}\text{D}1$\\
$\text{Res}{-}3{\times}3{-}\text{F}256{-}\text{D}2$\\
$\text{Res}{-}3{\times}3{-}\text{F}256{-}\text{D}2$\\
$\text{Res}{-}3{\times}3{-}\text{F}512{-}\text{D}4$\\
$\text{Res}{-}3{\times}3{-}\text{F}512{-}\text{D}4$\\
$\text{Conv}{-}3{\times}3{-}\text{F}512{-}\text{D}2$\\
$\text{Conv}{-}3{\times}3{-}\text{F}512{-}\text{D}2$\\
$\text{Conv}{-}3{\times}3{-}\text{F}512{-}\text{D}1$\\
$\text{Conv}{-}3{\times}3{-}\text{F}512{-}\text{D}1$\\

Where,
\begin{itemize}
  \setlength{\itemsep}{1pt}
  \setlength{\parskip}{0pt}
  \setlength{\parsep}{0pt}
	\item `Conv' stands for a convolutional layer, with ReLU activation~\cite{Nair10} and batch-normalisation~\cite{Ioffe15}; `Res' stands for the Pre-activation Residual Unit of He~\etal~\cite{he2016identity}.
	\item second term is dimensions of the the filters
	\item  `F$n$' means $n$ filters
	\item `D$r$' gives the dilation rate of the filters~\cite{YuKoltun2016}.
	\item if present, `S$2$' means a filter stride of 2, otherwise the stride is 1.
\end{itemize}

\subsection{Visual Object Counting Encoder}
The image encoder employed for counting is much simpler, and
employs six residual~\cite{he2016identity} layers. The image is not downsampled.
Layer-wise details, using the naming scheme given above, are as following:\\

\noindent $\text{Res}{-}5{\times}5{-}\text{F}32{-}\text{D}1{-}\text{S}1$\\
$\text{Res}{-}5{\times}5{-}\text{F}32{-}\text{D}1{-}\text{S}1$\\
$\text{Res}{-}5{\times}5{-}\text{F}32{-}\text{D}2{-}\text{S}1$\\
$\text{Res}{-}5{\times}5{-}\text{F}32{-}\text{D}2{-}\text{S}1$\\
$\text{Res}{-}5{\times}5{-}\text{F}32{-}\text{D}4{-}\text{S}1$\\
$\text{Res}{-}5{\times}5{-}\text{F}32{-}\text{D}4{-}\text{S}1$\\

\section{Recognising Multiple Lines of Text: Toy Example} \label{s:toy}
\begin{figure*}[t]
  \begin{minipage}[t]{\linewidth}
    \centering
    \includegraphics[width=\linewidth]{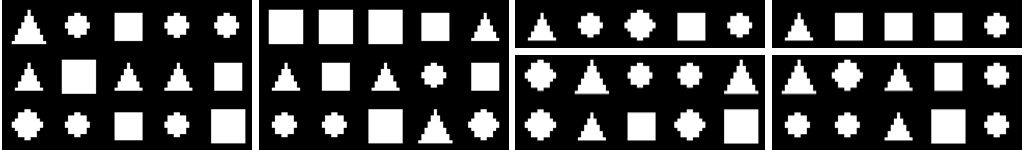}
  \end{minipage}\\

  \begin{minipage}[b]{\linewidth}
    \centering%
    \resizebox{\linewidth}{!}{%
    \begin{tabular}{ lrrrrrrrrrrrr }%
    \toprule
      $\downarrow$ Model {\textbackslash} Lines$\rightarrow$ & 1 & 2 & 3 & 4 & 5 & 6 & 7 & 8 & 9 & 10 & 15 & 20\\
      \midrule%
      end-to-end & 241.80 & 54.59 & 0 & 28.41 & 42.66 & 52.44 & 59.17 & 64.20 & 68.17 & 71.28 & 80.9 & 85.71\\
      inductive       & 0 & 0 & 0 & 0 & 0 & 0 & 0 & 0 & 0 & 0 & 0 & 0.02\\
      \bottomrule
    \end{tabular}}
    \par\vspace{0pt}
  \end{minipage}
  \vskip -0.1in
\caption{(top)
 Samples with different number of text lines from the Toy Shapes test set. (bottom) Normalised edit-distance rates (\%) for the task of recognizing shapes organised in blocks, comparing the generalisation capabilities of a conventional soft-attention RNN trained end-to-end, and with our inductive factorisation. The models were trained on blocks containing three lines, and tested for generalisation on a varying number of lines. Error rates of more than 100\% are due to the model always predicting exactly three lines.}
\label{fig:toy}
\end{figure*}

We evaluate generation ability of sequence recognition models to multiple lines on a synthetic Shapes dataset similar to the Coloured Shapes dataset used for counting. We experimented with the toy task of recognising sequences of shapes organised in multiple lines in an image (see~\cref{fig:toy}). Two models are evaluated: first, our inductive block parser, which is trained at the line-level. The other, a conventional soft-attention encoder/decoder RNN trained at the block level (no factorisation in terms of lines).

\paragraph{Dataset.} Binary valued images containing three types of shapes --- square, triangle, and circle --- each in two different sizes, and organised into lines, are synthetically generated. Each line consists of a sequence of five randomly sampled shapes. The training set consists of 2000 such images all containing \emph{three} lines. The test set consists of 12 different subsets with varying number of lines --- \{1 to 10, 15, 20\}, each containing 200 samples.
\paragraph{Evaluation.} We evaluate the models on images containing differing number of lines \{1--10,15,20\}, to test for generalisation. We report the normalised edit-distance, computed as the total edit-distance between the predicted block-string, and the ground-truth block-string, normalised by the length of the ground-truth string.
\paragraph{Model Architecture.} The two models, our inductive block parser and a conventional RNN model have the identical architectures: the image-encoder is a stack of six convolutional layers+ReLU (with $4{\times}16$, $2{\times}32$ filters, two $2{\times2}$ max-pooling after the second and the fourth-layers); the decoder is a soft-attention LSTM-RNN with 128 hidden units; attention embedding is also 128 dimensional. The memory updates $\Delta\bmm_t$ are regressed using two convolutional+ReLU layers (32 filters each).
\paragraph{Discussion.} The results are shown in \cref{fig:toy}. Both the models achieve perfect recognition accuracy on the test set containing three lines (the same number of lines as in the training set), but only the inductive line parser is able to generalise to different number of lines. The conventional RNN trained end-to-end to produce block-level predictions, always predicts three lines regardless of the number of lines in the test image.
\clearpage
\end{document}